# New design of Robotics Remote lab

Mohammad Alkafagee, Seifedine Kadry

Art, Sciences and Technology University, Lebanon
Email: skadry@gmail.com

**Abstract**

The Robotic Remote Laboratory controls the Robot labs via the Internet and applies the Robot experiment in easy and advanced way. If we want to enhance the RRL system, we must study requirements of the Robot experiment in a deeply way. One of key requirements of the Robot experiment is the Control algorithm that includes all important activities to affect the Robot; one of them relates the path or obstacle. Our goal is to produce a new design of the RRL includes a new treatment to the Control algorithm depends on isolation one of the Control algorithm's activities that relates the paths in a separated algorithm, i.e., design the (Path planning algorithm) is independent of the original Control algorithm. This aim can be achieved by depending on the light to produce the Light obstacle. To apply the Light obstacle, we need to hardware (Light control server and Light arms) and soft ware (path planning algorithm).The NXT 2.0 Robot will sense the Light obstacle depending on the Light sensor of it. The new design has two servers, one for the path (Light control server) and other for the other activities of the Control algorithm (Robot control server).The website of the new design includes three main parts (Lab Reservation, Open Lab, Download Simulation).We proposed a set of scenarios for organizing the reservation of the Remote Lab. Additionally, we developed an appropriate software to simulate the Robot and to practice it before usage the Remote lab.

## 1. Introduction

Since, the Control algorithm which includes all important activities to effect the Robot is one of key requirements of the Robot experiment, researchers should focus their efforts in developing new ways in design the Control algorithm to produce a new design of the RRL is more reliable, faster and stronger, so, one of most important questions will be asked here, how can we design the Control algorithm, (if we decide to replace or treat physical obstacles)? [3]. The answer is producing a new design of the RRL includes a new treatment to the Control algorithm depends on isolation one of the Control algorithm's activities that relates the paths or obstacles in a separated algorithm, i.e., design a (Path planning algorithm) is independent of the original Control algorithm. The Path planning algorithm will be responsible of planning the path only. The proposed solution will depend on the light to produce the path or obstacle, i.e., produce a Light obstacle can be sensed depending on the Light sensor of the NXT 2.0 Robot. Another question must ask here, how can we create the Light obstacle? The answer is by use a special hardware such as (Light control server and Light-arms) and a special soft ware such as the Path planning algorithm. The Light control server is responsible of control the Light-arms for lighting on a special area of Remote lab depending on orders of the Path planning algorithm to produce the Light obstacle, as we see in fig. 1.The Matlab can be used to design the Path planning algorithm.

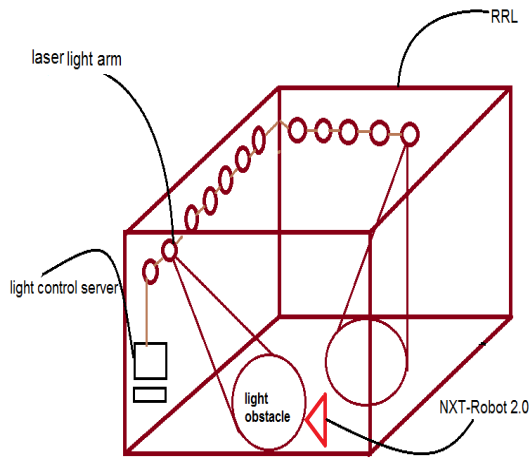

**Fig. 1: Generating the Light obstacle**

## 2. Background

In 1991, the first proposal for the Robotic Remote lab was presented. In 1992, testing first project of the Robot lab over the WAN [1]. In 1994, the first successful implementation of Robots lab via the Internet was developed by Goldberg at the University of South California. In 1995, McKee and Barson enhanced Remote lab by allowing Robot and its sensory devices in the lab to be controlled remotely. In 1996, the Rls were being updated by increasing the interactivity between users and RL. In 1997, updating automated measurement system of the RLs, to allow multiple users. In 1998, start with use a self contained camera system that can be controlled via the Internet. In 1999, Virtual laboratory was produced by the National University of Singapore (NUS). In 2002, start with use the RL in universities, such as University of South Australia that depends the RL in the lectures of it. In 2003, Automatic Control Telelab ACT system was produced at University of Siena. In 2009, the RLs became sharable among many users to treat the lack of modern labs.

## 3. Robotic Remote Laboratory

Robotic Remote laboratory (RRL) means control the Robot labs via the Internet, i.e., the Robot experiment which is running locally but directing externally. A user can be any computer connected to the Internet using a web browser and has the ability for monitoring and control the labs. This technique resulted from merging two spread fields (remote operations and the Robot labs), this merge, led to revolution in computer science and communication field, and find out a strong relationship between them. Remote laboratory makes the Robot experiment available 24 a day-7 days a week for any authorized user, i.e., sharing the lab. As well as, it provides a short cut access to the Robot experiment, i.e., each student or user can access the lab from any point has a connection to the Internet. When, we talk about Remote laboratory system, we must refer to topology of it, that consists of the Client side (users), and the Server side which has more than server (Robot server, Camera server) those deal directly with web server via the Internet. Many applications can appear with the Robotic Remote Laboratory, such as, Tele-teaching, Tele-maintenance, Tele-experiments, and Tele-production. To declare Remote laboratory in a best manner, we must distinguish between it and (Virtual laboratory), where, a user can interact with physical experiments instead of dealing with a graphical interface designed in software to simulate the reality, as well as, design and implementation of Remote lab are more complex and cost than Virtual lab.

## 4. Need for Robotic Remote Lab

The first purpose of using the Remote Lego laboratory is to allow students to compete remotely. The second is to treat lack of the modern laboratory in scientific institutes, by sharing the Robot labs among them. The third is to make Robot lab available 24 a day -7 days a week, for authorized users, no time limit. The forth is to enable users to apply their Robot experiment in easy and advanced way.

## 5. The proposed design

In this section we will propose a new design of the Robotic Remote Lab assist in improving the interactivity between users and the RRL and provide a large space of flexibility especially in the operation which relates constructing the remote path or (Obstacle). Whenever user wants to test his/her Robot experiment on the Remote Lab, he/she must request the website of the RRL that will be responsible of offering the interaction between remote user and the RRL. The Internet is a most suitable transfer for applying that purpose, the following figure will show the basic design for it.

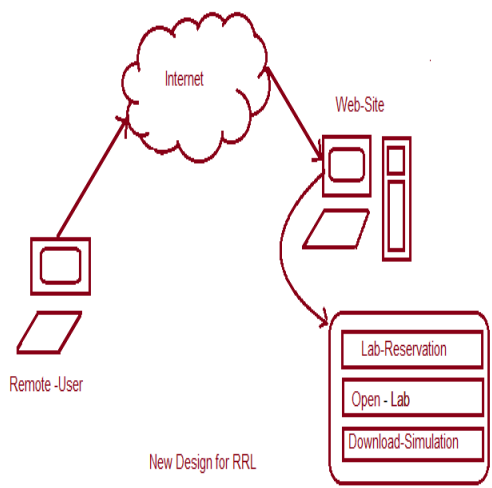

**Fig. 2: New design of the RRL**

In our proposed design the website of the RRL will include three main parts:

**1. Lab Reservation**: Which is responsible of organizing access to the RRL considering it as a critical recourse must be synchronized.

2. **Open Lab**: Which can access users to the RRL and this will not be discussed here.

3. **Download Simulation**: An application program which will simulate the reality of operations, we tend to apply it because the high cost of the Robots and the long distance of the Robotic Remote Lab.

If we consider our lab and software include 2 Robots both want to use same lab for competion, then, they shall either download the designed simulation for practicing their Robots in a simulating way by choosing the Download Simulation option, or choose the Open Lab option which enable them to access the Robotic Remote lab. Additionally, we can add an option to our software for generating the drawn path, and then to apply it physically to the lab by depending on the light techniques, i.e., we will provide the Remote lab with a further hardware for applying this purpose such as the Light control server and the Light-arms. As well as that, we will use a special input at user side such as the Mouse or the Touch Screen to apply the remote path. By mean of, whenever a user draws the path by the Mouse, the effect will be transferred to the Light control server of the Remote lab to generate the Light obstacles. The very important question must be asked here, how can we organize the reservation of the Remote Lab, if we take in consideration, the Remote Lab is a critical area must be synchronized ? To answer that, we suggest the following **scenarios** for dealing one or more users**:**

1.For **single user** option: A user has to create a new account and specify the date/duration for usage, after, the system will generate a pin and set it to that user, this pin will be valid only for the specified period. After the user log in, no other user can log in to the lab. The very important question must be asked here, how can we behave if two users want to access the same lab at the same time exactly? The answer is depending precedence for each user, i.e., a user which wasn't accessing that lab since last 24 hour will have a high primacy for access it. The lab "Open" option will be available after the specified period.

2. for **two users** option: A Coach ($3^{rd}$ person) should log in, create a new competition and specify the date/duration.
The system will generate 3 different pins, one for the Coach (Coach Privilege) and two pins (user privilege) for users.
During the competition, the Coach can start the competion, watch it online and declare a winner of the match. Any user wins in 4 matches will have the ability to be the Coach of the next match. If a user lost more than 5 matches, the system will recognize him/her and prevent him/her from the competion for the next 24 hour. Further, the system will prepare a suitable plan for training the losers.

3. for **three user's** option: First user should have a (high priority), second user should have a (middle priority) and third user should have a (low priority), each user has to create a new account for usage. The lab system will generate **3** different pins, (high privilege) for the first, (middle privilege) for the second and (low privilege) for the third, the lab system will prefer the first more than the others, i.e., the first can specify a (*t*) duration while the second and the third can specify a (*t/2*) and a (*t/3*) duration for usage. The pins will be valid only for the specified period. The lab will determine a number of users (only three) for each duration, so, after the three users log in, no other user can log in to the lab, the lab "Open" option will be available after the specified period.

4. For **four users or more** option: First user should be master and the others should be slaves, in this case, the master user can consider as a central controller for the slave users, i.e., if the master user follow a special path or perform any operation, the slave users must achieve that also. Besides, the master can specify the date/duration and determine a number of the slaves (at least three). We note here, in spite of applying the same operation by the slave Robots, the amount of speed for each slave Robot will depend on a user of it. The lab system will generate **four or more** different pins, one for the master user (master privilege) and three or more pins (slave privilege) for the slave users. We note, after the four users log in, no other user can log in to the lab; the lab "Open" option will be available after the specified period.

5. For **five users** option: The lab system will determine a number of users (only five) for each duration, deal with them at the same priority, no one better than one ( round table), in period way, each user will have a (time slice) for operation, i.e., share the duration time of the lab among users. All users shall be given the same date/duration for usage and each user has to create a new account. The system will generate **five** different pins; one pin for each user, this pin will be valid only for the specified period. After the five users log in, no other user can log in. The lab "Open" option will be available after the specified period.

6. for **six users or more** option: Parents (early two users) should login, specify the date/duration and determine a number of children (at least four). If one of parents want to log in alone, the system will refuse him/her and request two parents for trust, i.e., single user is not accepted, just for families, as we see in fig. 3. After the children log in, no other user can log in to the lab. The system will generate six or more different pins, two pins for the parents (parent privilege) and four or more pins (child privilege) for the children. These pins will be valid only for the specified period. The lab "Open" option will be available after the specified period. After the suggested scenarios, we can construct an obvious idea about the nature of the new design. The application program which will simulate the reality of operations will be used for practicing Robots before usage the Remote lab. We design and implement it because the high cost of the Robots and the long distance of the Robotic Remote Lab. the next chapter will produce appropriate discussion for third option of the new design which is Download Simulation.

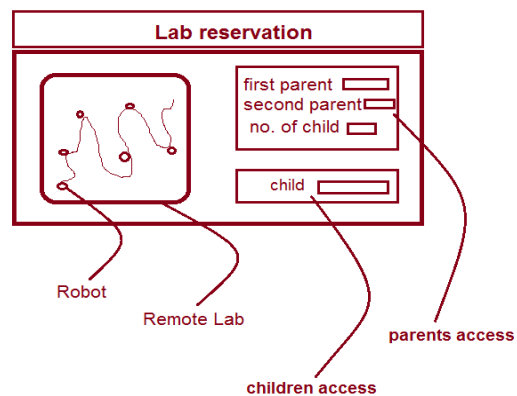

**Fig. 3: The Sixth scenario of lab reservation**

### 6. Simulation and Analysis

The goal of this section is to design and implement the third option of our new design which is **Download-simulation** that aims to simulate the operations of the Robotic Remote Lab and provide a simulation environment can be used to practice users before usage the remote lab. We do this simulation, because the high cost of the Robots and the long distance of the Robotic Remote Lab. Before start design and implement this application, we will take two case studies for the system, the first before applying the new design and the other after applying it to declare the new design in best manner as follows:

In the **first case**, a user designs the Control algorithm that contains all activities of the Robot as (path planning, speed determination,… etc.), then, a user sends the designed algorithm to the Robot control server to start the interaction between a user and the Robot. We see, all the activities of the Robot will be combined in one algorithm and send to one server for treating, as in fig. 4, if the server fails, the interaction will fail (no connection). As well as, the Robot control server must have a large memory to contain the algorithm that includes all activities. The (defined, predefined) path will be sent through the (designed, predesigned) Control algorithm to the Robot control server.

In the **second case**, a user designs path of the Robot alone in a special algorithm (Path planning algorithm) and the other activities alone in the (Control algorithm), then, the path algorithm will be sent to the Light control server to apply it, and the algorithm of the other activates to the Robot control server to perform it, if one of the servers fail, the activity of that server will fail, i.e., when the Light control server fail, the path activity will fail also, while the other activities will continue. We see here, the Control algorithm will be divided into two parts, the first is responsible of the path and the second is responsible of the other activities such as (speed, start, track,….etc.) as we see in fig. 5, each part has it's server. The system will become robust, since activities will be distributed between two servers instead of one. We see also, two operations shall be applied at the same time instead of one led to fast system.

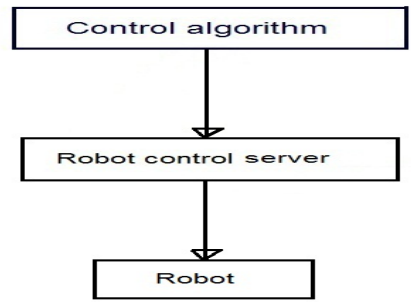

**Fig. 4 case study1**

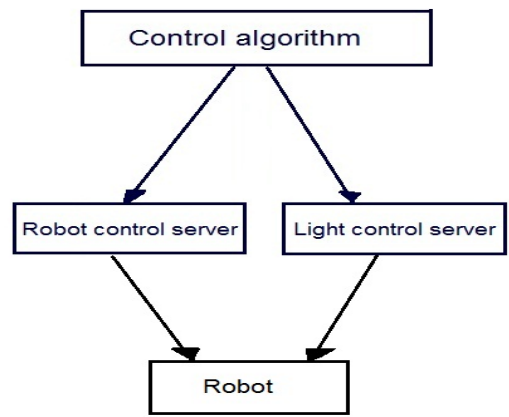

**Fig. 5 case study2**

At the first of the simulating program, we will start with predefine all important variables which will be used later, we see, most of them defied as a (Public) to make them usable by any part or object in the program considering them as a (global variables). We see also, a previous linking with the important libraries such as the sound-library (SpeechLib) and the graph-library (Graphics) which will be used as an important part in shaping colored lines, arcs and zigzag paths. Further, we determine type of color (red, blue), dimensions of arrays that uses for keeping track of the paths and set of variables which will be used in different parts in the program, as we see in the following orders**:**

```
PrivategAsGraphics
Dimp1AsNewPen(Color.Red)
Dim mrk
Dim spAsSpeechLib.ISpeechAudioPublic
path1(200, 2) As Integer Public i2 As Integer
```

Besides, we use the (Visual basic 2008) language as a programming tool offers an appropriate environment for simulation, the DELL-laptop-Studio 32- OS Intel(R) Core (TM) 2 Due CPU T5800 2.00GHZ 2.00GHZ (RAM) 4.00 GB and Windows 7 Ultimate.

**The Download Simulation** will be separated into two main parts:

1. Control Algorithm
2. Robot's behavior

### 6.1. Control Algorithm

The web site of our new design must provide the tools which are responsible of simulation the Control algorithm in easy and clear way. The behavior of the Robot will be changed according to change in the Control algorithm. The important activates that affect the Robot's behavior are the path, the speed, the track and the start/stop operation.

### A) Path Construction

This effect is responsible of simulation the path planning activity of the Control algorithm, offering the manner will be used for constructing the colored path. The Mouse is used to enter the path, so, the very important question here, how can we take the coordinates of the Mouse pointer during the movement of it? The answer will be via the programming sentence:
*(Private Sub Form1_Click(By Val sender As Object, By Val e As MouseEventArgs)Handles Me. Click).*
Depending on the previous sentence, any click on the form will led to keep the coordinates of the Mouse pointer in the system variables (*e.x,e.y*), then, they shall pass to the function (draw) which is responsible of drawing any required shape or path, if we consider, the color of the shape is determined previously. Other important question must ask here, how can the Robot sense the path and recognize the color of it ? The answer is by making the Robot follow a special frequency of color more than others, i.e., first Robot will follow the red color and recognize it only, while, the second will follow other color such the blue, as we see in fig. 6.

The following code will perform this operation:

```
PrivateSubdraw(ByVal         xpo        As
Integer,ByVal ypo As Integer)
         If    cl1    =    1    Then
g.DrawLine(p1,xp1,yp1,xpo,ypo)
         xp1=xpo
          yp1=ypo
      EndIf
      If    cl2    =2    Then
g.DrawLine(p2,xp2,yp2,xpo,ypo)
         xp2=xp
         oyp2=ypo
      EndIf
  g.Flush()
EndSub.
```

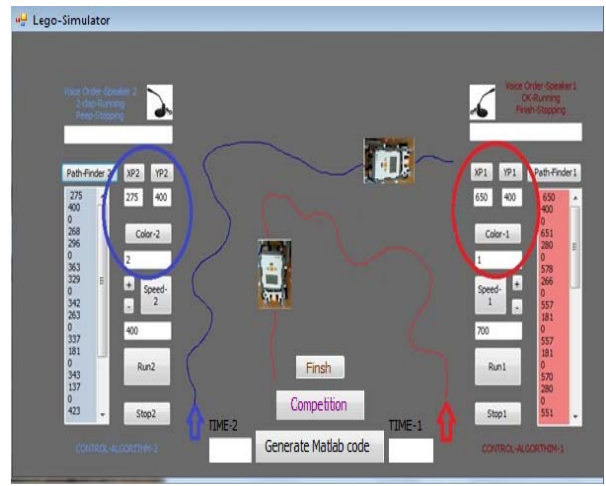

**Fig. 6: Path Construction by Mouse pointer**

### B) Speed Limitation

This effect is responsible of simulation the speed activity of the Control algorithm, offering tools for increase or decrease steps of the Robot. Using it, we can simulate the competion between two Robots. The time of each Robot will be computed during the match, as in fig. 7.

The following code will perform this operation:

```
Private Sub Button8_Click(ByVal sender
As System.Object)
    If   TextBox2.Text   =   1000   Then
MsgBox("You are accessed the limtation
of speed(High Speed!!!)")
         Timer1.Enabled = False
      ElseIf TextBox2.Text = 0 Then
MsgBox("You areaccessed the limtation
of speed(Low Speed!!!)")
         Timer1.Enabled    =    False
Else
Timer1.Interval=1000-Val(TextBox2.Text)
```

```
        End If
End Sub
```

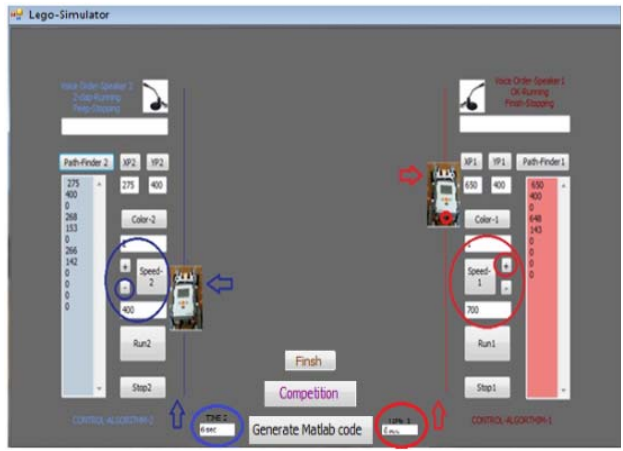

**Fig. 7: Using Speed limtations in competion**

### C) Run & Stop Operations

This effect is responsible of simulation the Start and Stop activity of the Control algorithm, offering two orders, the first can animate the Robot and the second can stop it.

There are two ways to apply these orders:

- **Mechanical order**: we can run the Robot by click on (**Run**) button, as we see in fig. 8, the following code will do this operation:
```
Private Sub Button5_Click(ByVal sender As
System.Object,ByValeAsSystem.EventArgs) HandlesButton5.Click
    If cl1=1 Then
        Timer1.Enabled=True
    EndIf
End Sub.
```
And we can stop it by clicking on (**Stop**) button, as we see in fig. 8.

The following code will do this operation:
```
Private Sub Button4_Click(ByVal sender As    System.Object,    ByVal    e AsSystem.EventArgs)HandlesButton4.Click
Timer1.Enabled = False
EndSub.
```
- **Vocal order :** in this case, we can run the Robot by giving a (**voice order**) to the Robot which can sense sound, for example, if we say the word (**ok**), the Robot will run and if we say the word (**finsh**) it will stop, as we see in fig. 8.

The following code will perform this operation:

```
PrivateSubTextBox3_TextChanged(ByValsenderAsSystem.Object,By
Val   e  As  System.EventArgs)  Handles TextBox3.TextChanged
    If TextBox3.Text <> "" Then
        IfTextBox3.Text="OK"Then
            If Timer1.Enabled =True Then
mrk=CreateObject("sapi.spvoice")mrk.speak
            ("Iam run now why you are repeated")
        Else Timer1.Start()
mrk=CreateObject("sapi.spvoice")mrk.speak("thank you for run me with order" & TextBox3.Text )
        End If
ElseIf
 TextBox3.Text = "Finish" Then
    If Timer1.Enabled = True Then
        Timer1.Stop()
mrk=CreateObject("sapi.spvoice")mrk.speak("why you are stopped me with order"& TextBox3.Text )
 Else
mrk = CreateObject("sapi.spvoice")
            mrk.speak("Iam  not running to stop me now ")
        End If
        End If
    End If
End Sub
```

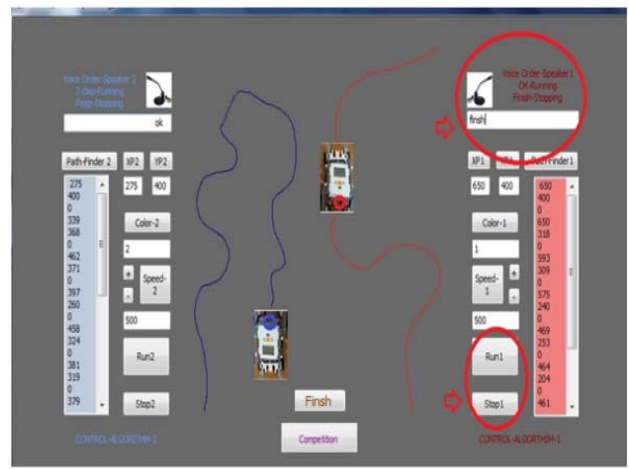

**Fig. 8: Using vocal and mechanical order to start and stop**

### D) Route Derivation:

This effect is responsible of simulation the (Keep track) activity of the Control algorithm, producing a final report of points which the Robot pass them through the travel of it, as we see in fig. 9.We can do that, by click on the (**Path-Finder**) button. The following code will perform this operation:

```
Private Sub chkpath1()
 n1 = n1 + 1
 path1(n1, 0) = xg
 path1(n1, 1) = yg
 End Sub
Private Sub prinpath1()
        TextBox7.Text = " "
        For i = 0 To n1 + 1
            For j = 0 To 2
                TextBox7.Text         =
TextBox7.Text & path1(i, j) & vbCrLf
            Next
        Next
    End Sub
```

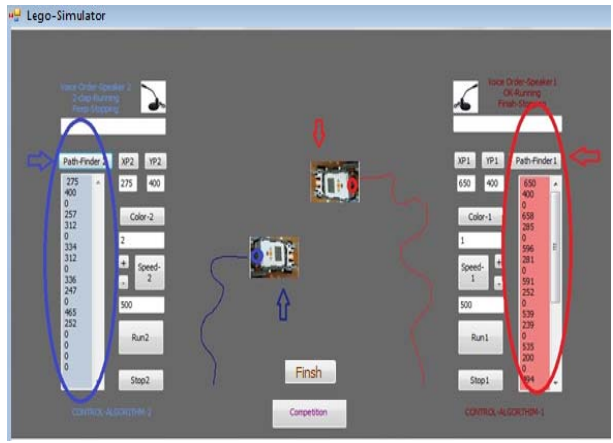

**Fig. 9: Final report of Route Derivation**

### 6.2. Robot's behavior

In this part we will simulate the relation between the Robot and the Control algorithm, so, we will see every effect in the Control algorithm will affect the Robot's behavior, i.e., draw the path, increase or decrease the speed, keep the track and start the operation, shall led to produce the final behavior of the Robot, such as select the correct path, rotate in a true angle and move to one of the four sides, as we see in fig. 10.

The following code will perform this operation:
```
 Private Sub Timer1_Tick(ByVal sender
As   System.Object,   ByVal   e   As
System.EventArgs) Handles Timer1.Tick
      If i2 Mod 2 = 0 Then

PictureBox1.Load("D:\games1\1.gif")
       Else

PictureBox1.Load("D:\games1\2.gif")
        End If
        i2 = i2 + 1
        My.Computer.Audio.Play(My.Resources.ss1
, AudioPlayMode.Background)
        If    PictureBox1.Location.Y    >
path1(n1, 1) Then
          PictureBox1.Top            =
PictureBox1.Top - 10
        Else
           If PictureBox1.Location.X >
path1(n1, 0) Then
pictureBox1.Image.RotateFlip(RotateFlip
Type.Rotate90FlipX)
            PictureBox1.Left           =
PictureBox1.Left - 5
          Else
       If   PictureBox1.Location.X    <
path1(n1, 0) Then
PictureBox1.Image.RotateFlip(RotateFlip
Type.Rotate90FlipY)
             PictureBox1.Left=
ictureBox1.Left + 5
             Else
PictureBox1.Image.RotateFlip(RotateFlip
Type.Rotate180FlipX)
               PictureBox1.Top     =
PictureBox1.Top + 10
            End If
          End If
       End If
    End Sub
```

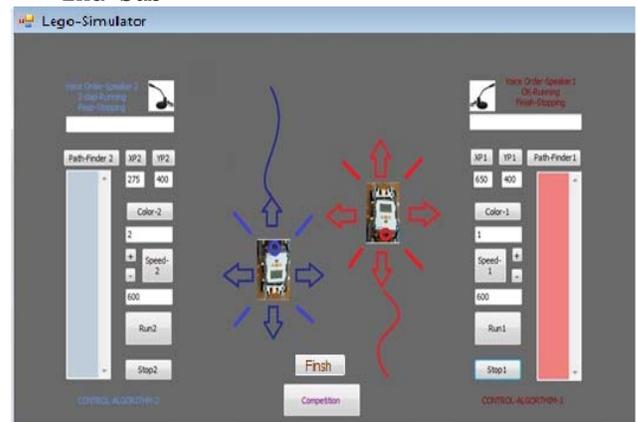

**Fig. 10: Robot's behavior**

### 7. Conclusion

After analyzing the results of the two case studies, we can conclude that applying the new design of the RRL will led to isolation the path planning activity of the Control algorithm in independent algorithm (path planning algorithm) depending on Light techniques, i.e., we provide another server for the RRL (Light control server) responsible of producing the path or the Light obstacle only, So, the new system will have two servers, one for applying the path(Light control server) and other for applying the other activates of the Control algorithm(Robot control server). The system will become stronger, faster and more reliable, since, the work is divided between two servers instead of one. As well as, the new design will open the door to huge developments in the RRL system depend on partition and distribution the other activities of the Control algorithm among many servers, each activity has it's server.

## 8. Future work

The future developments regard the possibility of use a **Virtual driver** for the Robot experiment**,** determine the speed of the Robot according to position of the Light obstacle, if, it is far, the Virtual driver will increase the speed (speeder), else it will decrease the speed to average, this feature will allow users to design and test speed change algorithm.